\documentclass{llncs}
\usepackage{amsmath}
\usepackage{graphicx}
\usepackage[nounderscore]{syntax}
\usepackage[linesnumbered,ruled]{algorithm2e}
\usepackage{xspace}

\newcommand{\Tree}{\emph{Tree}\xspace}
\newcommand{\Scenario}{\emph{Scenario}\xspace}
\newcommand{\Language}{\emph{Language}\xspace}
\newcommand{\Table}{\emph{Table}\xspace}

\begin{document}

\title{What is understandable in Bayesian network explanations?}

%
\titlerunning{}  
%
\author{Raphaela Butz\inst{1} \and Ren\'ee Schulz\inst{2} \and Arjen Hommersom\inst{1,3} \and Marko van Eekelen\inst{1,3}}
%
%
%
\institute{
Department of Computer Science, Open University of the Netherlands\\
\and
Department of Multimedia Engineering, Osaka University, Japan\\
\and
ICIS, Radboud University, The Netherlands}

\maketitle              

\begin{abstract}
Explaining predictions from Bayesian networks, for example to physicians, is non-trivial.
Various explanation methods for Bayesian network inference have appeared in literature, focusing
on different aspects of the underlying reasoning.
While there has been a lot of technical research, there is very little known about how well humans actually understand these explanations.
In this paper, we present ongoing research in which four different explanation approaches were compared through a survey by asking a group of human participants to interpret the explanations. 
\end{abstract}

\section{Introduction} \label{sec:into}
Bayesian networks (BNs) gained rising interest in AI in medicine in the last 30 years. Part of the interest in Bayesian networks in this particular domain lies in the apparent transparency and interpretability of models~\cite{arrieta2020explainable}. However, it has been widely recognised that explaining the inference of Bayesian networks is difficult as they do not imitate human reasoning \cite{lacave2002review}, which may hinder their applicability in the health care domain.
Therefore, there have been many proposals for explaining Bayesian networks in the field of healthcare, e.g.~\cite{butz18,Kyrimi:2016,Williams:2006} as well as in other fields \cite{Timmer:2017},\cite{Vlek:2016},\cite{Vreeswijk:2005},\cite{Yap:2008}. This shows that explaining BNs is not a trivial task and raises the question which methods or aspects of methods are actually useful in practice.

The goal of this paper is to compare and analyse the human understandability of BNs using recent explanation methods from Vlek~\cite{Vlek:2016}, Kyrimi~\cite{Kyrimi:2016}, Timmer~\cite{Timmer:2017} and Butz~\cite{butz18}, which are based on \cite{Williams:2006} \cite{Vreeswijk:2005} \cite{Yap:2008}. We first briefly introduce the explanation methods and our approach in Section~\ref{sec:methods}. We then discuss the survey that we used in Section~\ref{sec:case_study}. In Section~\ref{sec:evaluation} the results are evaluated and the conclusion of our findings can be found in Section~\ref{sec:conclusion and future work}.

\section{Methodology}\label{sec:methods}

For the survey, a mixed methods (qualitative and quantitative) approach was evaluated to be necessary. We require qualitative (e.g. opinions, questions, thoughts, and misunderstandings) as well as quantitative data (demographics, self evaluation of understanding, and comparison of models on a scale). Interviews as well as an online version of the interview in a survey format were conducted. The interview and survey feature the same set of questions and both methods are based on open-ended answer possibilities including a few scale-based and multiple choice questions.

Even though this research is done to improve the use of explained BNs in the medical and healthcare field, an approach was needed to broaden the range of possible participants. Therefore, we did not use a BN with medical data to minimize the necessity of background knowledge of the participants. Furthermore, the survey was designed using a gamified approach to ease the cognitive load on the participants' end. Gamification is defined as the use of game design elements used in non-game context \cite{deterding2011game}. In this survey, we chose the element of storytelling as well as contextualised (to the story) visualizations. The goal with this approach is to gain a generalisable outcome to improve the explanation of BNs in all contexts, including healthcare. 

For the analysis of the free text answers, the qualitative content analysis process \cite{Elo:2008} was chosen. The coding of the data was done with a deductive approach, meaning we chose analysing categories based on previous research. There were two sets of categories that were used for the analysis.
The first set of categories features the ability of participants to come to a conclusion based on the different explanation methods, with the subcategories of 'having doubts' or 'no doubts' about the conclusion. It should be noted that only the participants' perception and not their actual understanding of the methods were measured. The second set of categories describes the participants' expressions (positive and negative) about specific aspects of the explanation methods (e.g. certain visualization forms, structure or type of language). 
Furthermore, the coding of the data was done by two experts individually for comparable sets of coded data to ensure quality and further iteration of the categories as needed. The categories as stated in this paper were refined in 3 iterations, developed through the analysis of the content and deeper understanding of user behavior.

\section{Case Study and Survey Design}\label{sec:case_study}
For the design of the case study, several explanation methods were chosen which we labeled with the names \emph{Tree, Scenario, Table, Language}. The \Tree method by Timmer et al.~\cite{Timmer:2017} is using reasoning chains to explain the derivation of probabilities resulting from the BN. This method requires two phases. In the first phase, a support graph is constructed from the identified chains of reasoning featuring the target variable as a root node. In the second phase, arguments are subsequently constructed and presented in a (visual) tree structure for the users to interpret. 
\newline \indent Vlek et al.~\cite{Vlek:2016} \Scenario method uses Scenario nodes to evaluate the coherence of a scenario in a BN. Scenarios, as modeled in this method, add a narrative component to allow experts to gain overview over the entirety of the statistical information within a BN while being presented with smaller scenarios highlighting certain aspects of the BN's information. The scenario nodes in the BN allow for the calculation of how the evidence supports or refutes each scenario node. 
\newline \indent Kyrimi and Marsh~\cite{Kyrimi:2016} method's structure consists of three successive levels of detail featuring an incremental explanation. These levels are \emph{1: Significant evidence variables}, \emph{2: Information flow} and \emph{3: Significant evidence impact on the intermediate variables}. Level 1 determines in which way the evidence affects the variable of interest. Level 2 shows the intermediate variables through which the information passes in the BN. Level 3 shows in which way the evidence affects the variables through which the information passes.
\newline \indent The last method, \Language, is describing the most probable explanation in natural language and has been introduced in Butz, Hommersom and Van Eekelen~\cite{butz18}. A argumentative probability tree is built by searching for the strongest dependency starting from the evidence. To generate natural language from the argumentative probability tree, boiler plates were introduced and generated in a uniform structure due to the argumentative probability tree.
\newline \indent For all presented explanation methods, the BN fom Vlek~\cite{Vlek:2016} has been used as a basis to have a comparable set of explanations. This means that the same data was presented using the different explanation methods. During the design of the final representations of each method, we found that in some cases, creating such examples is not straightforward. 
\newline \indent Some of the explanation methods feature technical terminology of its own and therefore are not generally understandable without further explanation. One example for this can be found in \Table~\cite{Kyrimi:2016}, as the final representation features a certain structure of 'levels', however the potential reader of the results does not actually know what that means. This means that some methods generate another structure that is in need of explanation. To be able to include these methods in the comparison, we included a base set of explanations on how to read and understand certain parts of the representation, e.g. 'levels' in \Table and the indicator called 'infinite' in \Tree~\cite{Timmer:2017}. 
\newline \indent For the survey, we used Vlek's BN~\cite{Vlek:2016} on a homicide case to provide an equal data set in order to prevent the results from being biased by data sets of different quality. This was done with the intention that a non-specialized data set can be understood by everyone, without specialized knowledge, as it would have been necessary in the medical domain. An example for an explanation method based on a medical condition, that requires specialized knowledge to understand the explanation, can be found in \cite{Kyrimi:2016}. In the survey, we asked the participants to explain each method in a free text form or took notes while interviewing. At the end, we gave them the opportunity to reflect on the methods and to assess for themselves how informative they perceived them.

\section{Survey Results}\label{sec:evaluation}
In  total, 20 people from different professions participated in the survey. Out of these 20 participants, 3 claimed having basic knowledge about BNs, 9 heard of them, and 8 have never heard of them before. Participants indicated that the BN itself and \Scenario were the easiest to understand with 11 votes, while \Table was voted the most confusing with 11 votes followed by \Language with 8 votes. Likewise, 11 people stated that the BN gives the most insight, while 4 voted for \Scenario. \Language came in third with 3 votes. 

Some issues were frequently raised in some methods, such as missing explanations about the meaning of arrows and how visualisations were generated was asked 7 times. These issues were often mentioned in combination with missing explanations for colour coding (6 times) and the percentage indicators (9 times).
It was stated 5 times that in \Tree, the labeling with 'infinite' was confusing. Furthermore, the difference between the terms "infinite" and "certain" was not clear.
That \Table is confusing was stated 10 times, 5 times with a focus on the levels and 4 times the average scenario was not understood. 
That the language used in \Language was not natural and thus confusing was mentioned 6 times, whilst 4 times it was voted  the easiest to understand. 2 times, however, it was stated that the text was too long and that participants preferred information presented more to the point and shorter. 

\begin{figure}[h] 
\centering
\includegraphics[width=0.7\textwidth]{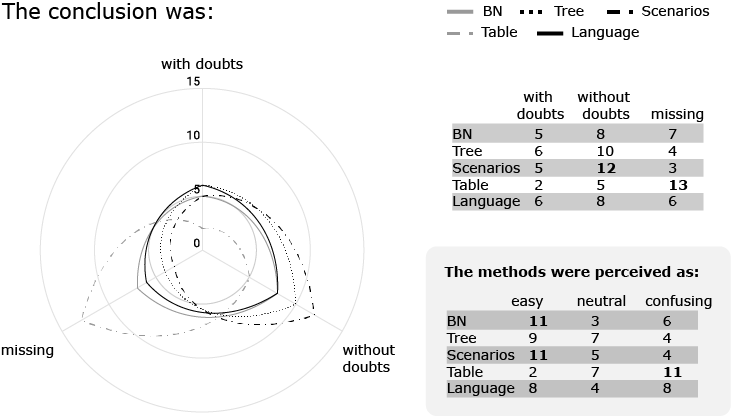}
\caption{Results of the qualitative content analysis by categories.}
\label{fig:survey}
\end{figure}

The results of the qualitative content analysis can be found in Figure~\ref{fig:survey}. The most prominent results are that the participants had the least doubts about their conclusion in \Scenario (12 people), whereas 13 people could not deduct a conclusion from \Table. Another interesting result is that the explanatory ability of the BN hardly differs from that of \Language in the presented example. 

\section{Conclusion}\label{sec:conclusion and future work}
A conclusion that can be drawn for all methods is that it is not only about developing a method to explain a BN, but it is also essential to state what prior knowledge is required to understand the explanation. The evaluation of \Table was particularly difficult for the authors because the paper did not specify what prior knowledge was necessary to understand the results of the method. Furthermore, it was noticeable that many people spoke of scenarios (13 times), not only in \Scenario, which suggests that we like to think in scenarios and want to see them confirmed or refuted.

Most participants stated \Scenario was the easiest to understand after the BN. This result may be biased because the example that was used in the case study was designed for the \Scenario method. Other types of data may have been more suitable for other explanation methods. 



In addition, it seems that a combination of visualization and simple, but precise text helps most participants to make decisions. Natural language (and longer notes or sentences) seems to be in a disadvantage for grasping information quick and precisely in a written format. However, as this research continues, we are aware that switching the output medium to e.g. audio might have further impact on the usefulness of a representation. 
\vspace{-0.5\baselineskip}

\bibliographystyle{splncs03}
\bibliography{bibliography} 

\end{document}